\documentclass[sigconf]{acmart}

\usepackage{booktabs}
\usepackage{balance} 
\usepackage{amsmath, amssymb, amsmath, mathtools, amsthm}
\usepackage[caption=false,font=footnotesize]{subfig}
\usepackage{graphicx, comment, epstopdf, multirow}
\usepackage{algorithm, ,algorithmicx}
\usepackage[noend]{algpseudocode} 
\usepackage{tikz, pgfplots, epstopdf, todonotes}
\usetikzlibrary{positioning}
\usepackage{colortbl}
\definecolor{Gray}{gray}{0.90}

\DeclareMathOperator*{\argmin}{argmin}

\newcommand{\FP}[0]{\ensuremath{F\!P}}
\newcommand{\FN}[0]{\ensuremath{F\!N}}
\newcommand{\vect}[1]{\ensuremath{\boldsymbol{#1}}}

\newcommand{\vm}[0]{\vect{m}}

\newcommand{\routePlan}[3]{P_{#1}\left({#3}_{#2}\right)}
\newcommand{\travelTime}[3]{T\left({#1}, {#3}_{#2}\right)}

\setcopyright{acmlicensed}

\acmDOI{http://dx.doi.org/10.1145/3063386.3063767}

\acmISBN{978-1-4503-4989-5/17/04}

\acmConference[SCOPE 2017]{The 2nd Workshop on Science of Smart City Operations and Platforms Engineering}{April 2017}{Pittsburgh, PA USA}
\acmYear{2017}
\copyrightyear{2017}
\acmPrice{15.00}

\begin{document}
\title[Optimal Detection of Faulty Traffic Sensors]{Optimal Detection of Faulty Traffic Sensors\\Used in Route Planning}

\author{Amin Ghafouri}
\orcid{1234-5678-9012}
\affiliation{%
  \institution{Vanderbilt University}
  \city{Nashville} 
  \state{Tennessee} 
  \postcode{37212}
}
\email{amin.ghafouri@vanderbilt.edu}

\author{Aron Laszka}
\affiliation{%
  \institution{Vanderbilt University}
  \city{Nashville} 
  \state{Tennessee} 
  \postcode{37212}
}
\email{aron.laszka@vanderbilt.edu}

\author{Abhishek Dubey}
\affiliation{%
  \institution{Vanderbilt University}
  \city{Nashville} 
  \state{Tennessee} 
  \postcode{37212}
}
\email{abhishek.dubey@vanderbilt.edu}

\author{Xenofon Koutsoukos}
\affiliation{%
  \institution{Vanderbilt University}
  \city{Nashville} 
  \state{Tennessee} 
  \postcode{37212}
}
\email{xenofon.koutsoukos@vanderbilt.edu}

\renewcommand{\shortauthors}{A. Ghafouri et al.}

\begin{abstract}
In a smart city, real-time traffic sensors may be deployed for various applications, such as route planning. Unfortunately, sensors are prone to failures, which result in erroneous traffic data. Erroneous data can adversely affect applications such as route planning, and can cause increased travel time. To minimize the impact of sensor failures, we must detect them promptly and accurately. However, typical detection algorithms may lead to a large number of false positives (i.e., false alarms) and false negatives (i.e., missed detections), which can result in suboptimal route planning. In this paper, we devise an effective detector for identifying faulty traffic sensors using a prediction model based on Gaussian Processes. Further, we present an approach for computing the optimal parameters of the detector which minimize losses due to false-positive and false-negative errors. We also characterize critical sensors, whose failure can have high impact on the route planning application. Finally, we implement our method and evaluate it numerically using a real-world dataset and the route planning platform OpenTripPlanner.
\end{abstract}

%
%



\maketitle

\section{Introduction}
 

In smart cities, real-time traffic sensors may be deployed for various applications. However, sensors are prone to failures, which result in erroneous traffic data. Erroneous data can adversely affect the performance of applications. 
To minimize the impact of sensor failures, we must detect them promptly and with high accuracy. However, typical detection algorithms may lead to a large number of false positives and false negatives, which can result in suboptimal performance. 

Anomaly detection of faulty traffic sensors has been studied in the literature. Typical approaches include using data-driven methods that incorporate historical and real-time data to detect anomalies \cite{lu2008faulty}, \cite{zygourastowards}, \cite{robinson2006development}, \cite{widhalm2011identifying}. However, existing approaches may result in high performance-losses in traffic applications, mainly due to false-positive (FP) and false-negative (FN) errors. 
In order to minimize the losses, it is desirable to reduce the FP and FN rates as much as possible. But, there exists a trade-off between them, which can be changed through a detection threshold.
To address this, it is necessary to 
take into account the traffic application when designing anomaly detectors, and quantify the losses in the traffic application caused by the FP and FN errors. By selecting the right detection threshold, the performance losses caused by FPs and FNs can be minimized.


In this paper, we study the problem of finding optimal thresholds for anomaly detection of faulty traffic sensors, considering route planning as the application of interest. The objective is to select the optimal thresholds of anomaly detectors in order to optimize the performance of the route planning application in the presence of faulty sensors. 
We devise an effective detector for identifying faulty traffic sensors using a prediction model based on Gaussian Processes. Further, we present an approach for computing the optimal parameters of the detector which minimize losses due to false-positive and false-negative errors.  
We also characterize critical sensors, whose failure can have high impact on the traffic application. 
Finally, we implement our method and evaluate it numerically using a real-world dataset and the route planning platform OpenTripPlanner \cite{mchugh2011opentripplanner}.
Our evaluation results show that the proposed strategy successfully minimizes the performance loss and identifies the critical sensors.

The remainder of this paper is organized as follows. In Section~2, we present the background for route planning and Gaussian Process regression. In Section~3, we introduce the system model. In Section~4, we define a notion of optimal detection, present a method to obtain near-optimal thresholds, and define critical sensors. In Section~5, we implement our method and evaluate it numerically. In Section~6, we discuss related work. Concluding remarks are presented in Section~7.

\section{Preliminaries}


\subsection{Route Planning}
Let $G = (V,E)$ be a directed graph with a set $V$ of vertices and a set $E$ of arcs.  Each arc $(u, v) \in E$ has an associated nonnegative cost $c(u, v)$. The cost (i.e., length) of a path is the sum of the costs of its arcs. In the point-to-point shortest path problem, one is given as input the graph $G$, a query $q=(o,d)$, where $o \in V$ is an origin and $d \in V$ is a destination, and the objective is to find a minimum-cost (i.e., shortest) path from $o$ to $d$ in~$G$. 
In the many-to-many shortest path problem, a set of queries $Q$ is given, and the goal is to find the minimum-cost path for each query $q = (o,d) \in Q$. 

There exist many route planning algorithms that compute optimal solutions in an efficient manner \cite{bast2016route}. Among these methods, the bidirectional Dijkstra's algorithm with binary heaps computes point-to-point shortest path in $\mathcal{O}(|E|+|V| \log |V|)$. Further, the Floyd-Warshall algorithm solves all pairs shortest paths in $\mathcal{O}(|V|^3)$. A large number of methods have been designed to improve running time of shortest-path algorithms. For example, contraction hierarchies and arc flags have been successfully used~\cite{dibbelt2015user}.

\subsection{Gaussian Process Regression}

GPs provide a Bayesian paradigm to learn an implicit functional relationship $y = f(\vect{x})$ from a training dataset  $\{(\vect{x}_i,y_i);i=1,2,...,n\}$, where $\vect{x}_i \in \mathbb{R}^d$ represents the vector of observed input variables (i.e., predictors), and $y_i$ is the observed target value. A comprehensive discussion of GPs in machine learning can be found in~\cite{rasmussen2006gaussian}.

GPs directly elicit a prior distribution on the function $f(\vect{x})$, and assume it to be a GP a priori,
\begin{equation}
f(\vect{x}) \sim GP\left(\mu(\vect{x}),k(\vect{x},\vect{x'})\right) .
\end{equation}
For a new point $\vect{x}_*$, the goal is to predict $y_* = f(\vect{x}_*)$. Given that the regression function is a GP, the distribution of the values of $f$ at any finite number of points is a multivariate Gaussian distribution. Therefore,
\begin{equation}
\begin{pmatrix} \vect{y} \\
y_* \end{pmatrix} \sim \mathcal{N}\Biggl(\mu(x),\begin{pmatrix} K & K'_*\\
K_* & K_{**} \end{pmatrix}\Biggr) ,
\end{equation}
where $K$ is the covariance matrix for the labeled points, $K_*$ is the covariance vector between the new point and the labeled points, and $K_{**}$ is the measurement noise.
Then,
\begin{equation}
\Pr(y_* \,|\, \vect{y}) \sim \mathcal{N}\left(K_*K^{-1}\vect{y},K_{**}-K_*K^{-1}K'_*\right) .
\end{equation}

The prediction of a GP model depends on the choice of covariance function, which identifies the expected correlation between the observed data. Typically, a parametric family of functions is used, and the hyperparameters are inferred from the data. Examples of the commonly used covariance functions include polynomial kernel, automatic relevance determination (ARD), and radial basis function (RBF). Methods for learning the hyperparameters are based on maximization of the marginal likelihood, which can be performed using gradient-based optimization algorithms. 

\section{System Model}
In this section, we present the system model. 
We first define a model of transportation network. Then, we construct a detector for identifying faulty traffic sensors using a prediction model based on Gaussian Processes.

\subsection{Transportation Network}
Consider a transportation network modeled as a graph $G=(V,E)$, where edges represent road segments and vertices represent connections between road segments (e.g., traffic junctions). We assume that a subset $S \subseteq E$ of the road segments are monitored by sensors that measure traffic state (e.g., speed, occupancy, flow) at discrete timesteps $k\in \mathbb{N}$. 
The measurements of these sensors are transmitted to a navigation service, which given a set of queries $Q(k)$ at timestep $k$, computes the corresponding shortest paths. For segments without a traffic sensor, we assume the navigation service uses either previously computed values or predicted values using measurements of adjacent sensors.

Traffic sensors may be faulty due to miscalibration or hardware failure. If a sensor $s \in S$ is faulty, there is a discrepancy between the actual and measured values. In other words, if $a_s(k)$ is the actual value and $m_s(k)$ is the measured value of faulty sensor $s$, then $m_s(k) = a_s(k) + \varepsilon_s(k)$, where $\varepsilon_s(k)$ is the fault value at time $k$. In this model, we do not consider faults that result in no data being sent, since such cases can easily be filtered out by an operator. 

\subsection{Gaussian Process-Based Detector}
Given the sensor measurements, we need to decide whether some sensors are faulty. 
We assume that the number of sensors that simultaneously become faulty is low, which is true in practice. As a result, for any sensor, the majority of nearby sensors that have not been marked faulty provide reliable traffic data, and so we can use these nearby sensors to predict the value measured by the sensor in question. To detect faults, we then compare the predictions to the measurements, and if there is a significant difference between the predicted values and the received measurements, an alarm indicating presence of a fault in that particular sensor is triggered.


\subsubsection{Traffic Prediction}
As our traffic predictor, we use GPs, which is a kernel-based machine learning method. Kernel-based methods have gained special attention for traffic prediction because of their generalization capability and superior nonlinear approximation. Among different kernel-based methods, previous work shows that GPs outperform other methods such as ARIMA and neural networks \cite{xie2010gaussian}. We use GPs because in addition to the above advantages, it allows for explicit probabilistic interpretation of forecasting outputs. 

As the kernel function, we decide for the commonly used ARD squared exponential,
\begin{equation}
K(\vect{m}(k),\vect{m}(k)')=\sigma_f^2 \exp \left(-\frac{1}{2}\sum_{i=1}^{d}\frac{(m_{i}(k)-m'_{i}(k))^2}{\sigma_i^2}\right) ,
\end{equation}
where $\vect{m}(k)$ and $\vect{m}(k)'$ are vectors of measurements, and $\sigma_f$ and $\{\sigma_i\}_{i=1}^{d}$ are hyperparameters. 



We let the target variable be the predicted traffic value $p_s$ (e.g., traffic flow or occupancy) of sensor $s \in S$ at timestep $k$. Further, we let the predictor variables be the measured traffic values of other sensors at the same timestep. In practice, two sensors are highly correlated if they are in close proximity. Therefore, it is possible to select predictor variables as the measured values of $d$ closest sensors from the target sensor, where the choice of $d$ depends on the network structure. This way, the predicted traffic value is defined as $p_s(k) = f(m_{V(s)}(k))$, where $V(s)$ is the set of $d$ closest sensors from $s$.

\subsubsection{Detection Algorithm}
We can efficiently detect failures for each sensor $s \in S$, by comparing the measured traffic value $m_s(k)$ with the predicted traffic value $p_s(k)$. We use Cumulative sum control chart (CUSUM) as the detection algorithm, which is a sequential analysis technique typically used for monitoring change detection \cite{page1954continuous}. 

Consider sensor $s \in S$, with a sequence of measurements \allowbreak$m_s(1), ..., m_s(k)$ and corresponding traffic predictions with means $p_s(1), ...,p_s(k)$ and standard deviations $\sigma_s(1), ..., \sigma_s(k)$. The standardized residual signal is defined as
\begin{equation}
z_s(k)= \frac{m_s(k)-p_s(k)}{\sigma_s(k)} .
\end{equation}
Moreover, upper and lower cumulative sums are defined as,
\begin{equation}
U_s(k) = \max (0,U_s(k-1)+z_s(k)-b_s) ,
\end{equation}
\begin{equation}
	L_s(k) = \min (0,L_s(k-1)+z_s(k)+b_s) ,
\end{equation}
where $U_s(k)=L_s(k)=0$ for $k=1$, and $b_s$ is a small constant.

Denoting the detection threshold at timestep $k$ by $\eta_s(k)$, a measurement sequence violates the CUSUM criterion at the sample $z_s(k)$ if it obeys $U_s(k) > \eta_s(k)$ or $L_s(k) < -\eta_s(k)$. Formally, letting $H_0$ and $H_1$ be the null and fault hypothesis, the decision rule is described by
\begin{equation}
d_s(U_s(k),L_s(k)) = \left\{ \begin{array}{lll}
H_1 & \textrm{  if } U_s(k)>\eta_s(k) \textrm{ or } L_s(k)<-\eta_s(k) \\ 
H_0 & \textrm{ otherwise} \\ 
\end{array}\right. .
\end{equation}

\subsubsection{False-Negative and False-Positive Trade-off}
In anomaly detectors, there might be a \textit{false negative}, which means failing to raise an alarm when a fault did happen.
Further, there might be a \textit{false positive}, which means raising an alarm when the sensor exhibits normal behavior.
It is desirable to reduce the FP and FN probabilities as much as possible. But, there exists a trade-off between them, which can be controlled by changing the threshold. In particular, by decreasing (increasing) the threshold, one can decrease (increase) the FN probability and increase (decrease) the FP probability.  

We represent the FN probability for each sensor $s$ by the function $\FN_s:~\mathbb{R}_+~\to~[0, 1]$, where $\FN_s(\eta_s(k))$ is the probability of FN when the threshold is $\eta_s(k)$, given that the sensor is faulty. Similarly, we denote the attainable FP probability for each sensor $s$ by $\FP_s:~\mathbb{R}_+ \to~[0,1]$, where $\FP_s(\eta_s(k))$ is the FP probability when the threshold is $\eta_s(k)$, given that the sensor is in normal operation. It is possible to plot the FP probability as a function of the FN probability for various threshold values~\cite{fawcett2006introduction} (e.g., see Figure~\ref{fig:roc}). 


\section{Optimal Detection}

In this section, we formulate the problem of finding optimal thresholds for anomaly detection of traffic sensors, considering route planning as their primary application. 
The objective is to select the optimal thresholds for anomaly detectors in order to minimize the losses caused by false positives and false negatives.
Then, we present an algorithm to find near-optimal detection thresholds. Finally, we characterize critical sensors, whose failure can have high impact on the traffic application.

\subsection{Optimization Problem}
First, consider the set of queries $Q$, and a route planning algorithm that takes as inputs the set of queries and the measured and predicted traffic values, and outputs the optimal routes. 
For a single query $q \in Q$ and sensor $s \in S$, we denote by $\routePlan{q}{s}{m}$ the optimal route computed using the measured traffic values for all sensors, and we denote by $\routePlan{q}{s}{p}$ the optimal route using the predicted value $p_s$ for sensor $s$ and the measured values $\vm_{-s}$ for all other sensors.
Finally, for a given route $r$ and sensor $s$, let $\travelTime{r}{s}{m}$ and $\travelTime{r}{s}{p}$ be the total travel time based on the measured $m_s$ and predicted $p_s$ values for sensor $s$, respectively, and the measured values $\vm_{-s}$ for all other sensors.

Then, $\travelTime{\routePlan{q}{s}{p}}{s}{m}$ is the measured travel time of the shortest route computed using the predicted value $p_s$ for sensor $s$. Similarly, $\travelTime{\routePlan{q}{s}{m}}{s}{m}$ is the measured travel time of the shortest route computed using the measured value~$m_s$. We define the loss caused by a false positive as follows:
\begin{equation}\label{cfp}
C_{s,q}^{\FP}(p_s, m_s) = \travelTime{\routePlan{q}{s}{p}}{s}{m} - \travelTime{\routePlan{q}{s}{m}}{s}{m} ,
\end{equation}
that is, the difference in measured travel time between using either the predicted or the measured value for sensor $s$.

The rationale behind the above expression is the following.
In case of a FP, according to the detector, the measured value $m_s$ is incorrect, but it is actually correct. Consequently, we choose a route that is computed using our prediction $p_s$ instead of the optimal route, which would be computed using the measurement $m_s$. To quantify the loss, we need to compare the travel times of the two routes, and we must use the measured traffic value $m_s$ for this comparison since that is the correct value in this case.

Similarly, for a FN, $\travelTime{\routePlan{q}{s}{m}}{s}{p}$ is the predicted travel time of the shortest route using measured value $m_s$, and $\travelTime{\routePlan{q}{s}{p}}{s}{p}$ is the predicted travel time of the shortest path using predicted value $p_s$. The loss caused by a FN is
\begin{equation}\label{cfn}
	C_{s,q}^{\FN}(p_s, m_s) = \travelTime{\routePlan{q}{s}{m}}{s}{p} - \travelTime{\routePlan{q}{s}{p}}{s}{p} ,
\end{equation}
that is, the difference in predicted travel time between using either the measured or the predicted value for sensor $s$. Note that in \eqref{cfp} and \eqref{cfn}, the values of $P$ and $T$ can be computed using existing route planning algorithms \cite{bast2016route}.

Next, let $\FP_s(\eta_s(k))$ and $\FN_s(\eta_s(k))$ be the probabilities of false-positive and false-negative errors when detection threshold $\eta_s(k)$ is selected. Further, let $p_f$ be the probability of fault, and let $p_n = 1-p_f$ be the probability of normal operation. For a given query $q$, the total loss caused by FPs and FNs is,
\begin{equation}\label{loss}
\begin{split}
L_{s,q}(\eta_s(k)) = & \FP_s(\eta_s(k)) \cdot C_{s,q}^{\FP}(p_s, m_s) \cdot p_n + \\ & \FN_s(\eta_s(k)) \cdot C_{s,q}^{\FN}(p_s, m_s)  \cdot p_f \; . 
\end{split}
\end{equation}

Considering the set of all queries $Q$, the total loss is 
\begin{equation}\label{lossall}
L_s(\eta_s(k), Q) = \sum_{q\in Q} L_{s,q}(\eta_s(k)) ,
\end{equation}
which allows us to define the notion of optimal detection threshold for a sensor.

\begin{definition}[Optimal Detection] The detection threshold~$\eta^*_s(k)$ is optimal for sensor $s$ if it minimizes the loss function~\eqref{lossall}. Formally, $\eta^*_s(k)$ is optimal for sensor $s$ if
	\begin{equation}\label{opt}
	\eta^*_s(k) \in \argmin_{\eta_s(k)} L_s(\eta_s(k), Q) .
	\end{equation}
\end{definition}

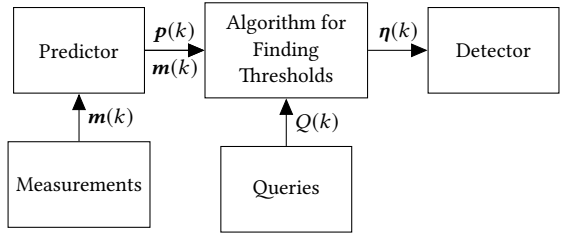
\begin{figure}
	\centering
	\begin{tikzpicture}[font=\small]
	\usetikzlibrary{arrows}
	\tikzstyle{box}=[draw, minimum width=5.5em, minimum height=3.6em];
	\tikzstyle{connector}=[->, >=triangle 45];
	
	\node [box] (predictor) {Predictor};
	\node [box, below =2em of predictor] (measurements) {Measurements};
	\node [box, right =2.5em of predictor] (algorithm) {\begin{tabular}{c}Algorithm for\\ Finding\\ Thresholds\end{tabular}};
	\node [box, below =2em of algorithm] (queries) {Queries};
	\node [box, right =2.5em of algorithm] (detector) {Detector};
	\draw [connector] (measurements) -- (predictor) node [midway, right] {$\vect{m}(k)$};
	\draw [connector] (predictor) -- (algorithm) node [midway, above ] {$\vect{p}(k)$};
	\draw [connector] (predictor) -- (algorithm) node [midway, below ] {$\vect{m}(k)$};
	\draw [connector] (queries) -- (algorithm) node [midway, right] {$Q(k)$};
	\draw [connector] (algorithm) -- (detector) node [midway, above] {$\vect{\eta}(k)$};
	\end{tikzpicture}
	\caption{Information flow in our approach.}
	\label{fig:diagram}
\end{figure}

Figure~\ref{fig:diagram} shows the flow of information in our approach. At each timestep $k$, given measurements $\vm(k)$, the predictor computes the predicted measurements $\vect{p}(k)$. Then, given a set of queries $Q(k)$, and the predictions and measurements, the thresholds $\vect{\eta}(k)$ are computed for the detectors using the algorithm presented next.




\subsection{Algorithm for Obtaining Thresholds}
We present Algorithm \ref{algo} to find near-optimal detection thresholds. The algorithm implements a random-restart hill climbing technique. If the FP to FN trade-off curve is convex, which makes \eqref{lossall} convex, we are able to compute optimal thresholds using convex optimization methods. However, this is not generally the case, as trade-off curves tend to be non-convex (see Figure \ref{fig:roc} for an instance of a trade-off curve).

The algorithm considers each sensor separately, and finds its corresponding detection threshold. At each iteration, the algorithm selects a new starting point and finds a local minimum using gradient-based optimization. In order to avoid unnecessary computation, we skip computing detection thresholds for sensors with very similar measured and predicted traffic values. Formally, for sensor $s \in E$, we select detection threshold $\eta_s = \infty$, if $|z_s(k)| < b$. This is because the detector's statistics $U_s(k)$ and $L_s(k)$ are decreasing and it is unlikely that an alert would be raised if one was not raised before. 


\begin{algorithm}[h!]
	\caption{Algorithm for Obtaining Thresholds}
	\label{algo}
	\begin{algorithmic}[1]
		\State \textbf{Input} $Q$, $\vect{\FP}(\eta)$, $\vect{\FN}(\eta)$, $\alpha$, $\gamma$
		\State \textbf{Initialize:} $\vect{\eta} \gets \vect{\eta}_0$, $\vect{L} ^\ast \gets \infty$
		\ForAll {$s\in S$}
		\If {$|z(k)| \leq b$ }
		\State  $\eta_s^\ast \gets \infty$ 
		\Else
		\While {$i < N$}
		\State $\eta_{s, new} \hookleftarrow \FP_s^{-1}($Uniform$([0,1]))$
		\State $\eta_{s, old} \gets 0$ 
		\thickmuskip=1mu
		\While {$|L_s(\eta_{s ,new},Q) - L_s(\eta_{s,old},Q)| > \alpha$}
		\thickmuskip=5mu plus 5mu
		\State $\eta_{s, old} \gets \eta_{s, new}$
		\State $\eta_{s, new} \gets \eta_{s, old} - \gamma \nabla_{\eta_s} L_s(\eta_{s,old},Q)$
		\EndWhile
		\If {$L_s(Q,\eta_{s,new}) < L_s^\ast$}
		\State $\eta_s^\ast \gets \eta_{s,new}$
		\State $L_s^\ast \gets L_s(\eta_{s,new},Q)$
		\EndIf
		\State $i \gets i+1$
		\EndWhile
		\EndIf
		\EndFor
		\State \textbf{return} $\vect{\eta}^*$
	\end{algorithmic}
\end{algorithm}

\subsection{Critical Sensors}
Value of the optimal loss gives insight on the criticality of traffic sensors. Fault on a sensor that has high loss value degrades the system's performance more than fault on a sensor with low loss value. We formally define the set of $\delta$-critical sensors below.
\begin{definition}[Critical Sensors] Set of $\delta$-critical sensors in a time period $[1, T]$ is defined as the set of sensors which have the average optimal loss values of greater than or equal to $\delta$. That is to say, a sensor $s$ is critical if $\frac{1}{T}\sum_{k=1}^{T} L_s(\eta^*_s(k),Q(k)) \geq \delta$.	
\end{definition}

Identifying critical sensors is beneficial, since it allows us to locate the most vulnerable elements of a network, which should be strengthened first to increase the robustness of a network. For example, if we have a limited budget which permits us to replace only a subset of the sensors with more robust ones, then we should start with the critical sensors.


\section{Evaluation}

In this section, we implement our method and evaluate it numerically using a route planning platform.

\subsection{System Model}
\subsubsection{Traffic Data} 
\begin{figure}
	\centering
	\includegraphics[width=.85\linewidth]{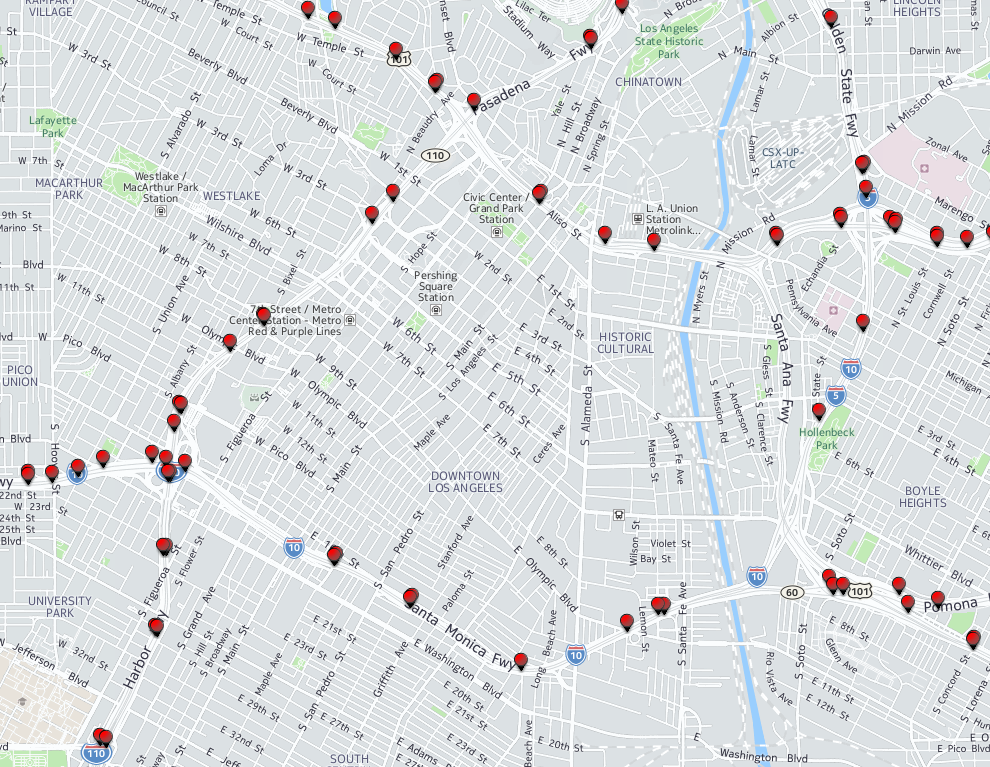}
	\caption{A map of traffic sensors installed in Downtown Los Angeles.}
	\label{fig:fig-case}
\end{figure}

We use a traffic dataset obtained from the Caltrans Performance Measurement System (PeMS) database \cite{caltrans}. 
The database provides real-time and historical traffic data from over 39,000 individual sensors, which span the freeway system across metropolitan areas of the State of California. 
Figure~\ref{fig:fig-case} shows the location of sensors in our case study, in which a total of 40 sensors are considered. We use the 5-minute aggregated data collected on the weekdays of September 3, 2016 to September 17, 2016. The dataset contains 115,200 data points. The first 7 days are used as training data, and the remaining 7 days are used as test data. 

To simulate faults, we use models for a specific set of fault types and ranges of fault magnitudes, which is similar to the approach presented in \cite{widhalm2011identifying}. The fault models are:
1) Constant Relative Overcount (caused by e.g., unsuitable sensitivity levels); range: $3\%$ to $7\%$ of the actual values (i.e., $\varepsilon_s(k) = u_s a_s(k)$ where $0.03 \leq u_s \leq 0.07$),
2)~Conditional Undercount (caused by e.g., sensor saturation); range:~$7\%$ to $13\%$ (i.e., $\varepsilon_s(k) = u_s a_s(k)$ where $-0.13 \leq u_s \leq -0.07$).

\begin{figure}
	\centering
	\includegraphics[width=.85\linewidth]{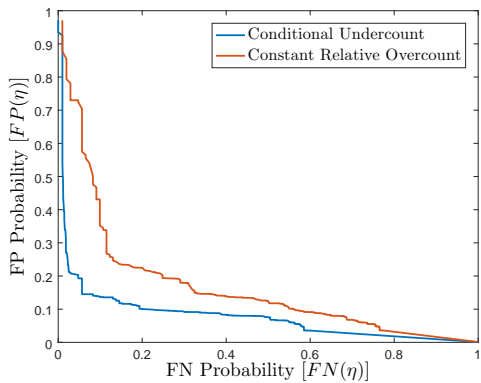}
	\caption{Trade-off between the false-positive and false-negative probabilities.}
	\label{fig:roc}
\end{figure}

Next, for each sensor, we construct a predictor using the measurements of its $d$ closest sensors as the predictor variables. We select $d=10$ since it results in the minimum overall prediction error.
We choose $b_s=0.05$ for all the detectors, to make them sensitive to small shifts in the mean. 
We evaluate each detector's performance by plotting the FP probability against the FN probability at various threshold values. Figure \ref{fig:roc} shows the trade-off curve of the detector implemented for a sensor, whose identifier in the PeMS dataset is VDS 774685. 

\subsubsection{Route Planner}
We use OpenTripPlanner (OTP), which is an open source platform for multi-modal route planning \cite{mchugh2011opentripplanner}. 
OTP relies on open data standards including OpenStreetMap for street networks. 
The default routing algorithm in OTP is the $A^*$ algorithm with a cost-heuristic to prune the search. For improved performance on large networks, it also uses contraction hierarchies. 

\subsection{Results}
\begin{figure}
\centering
\subfloat[]{\includegraphics[width=2.1in]{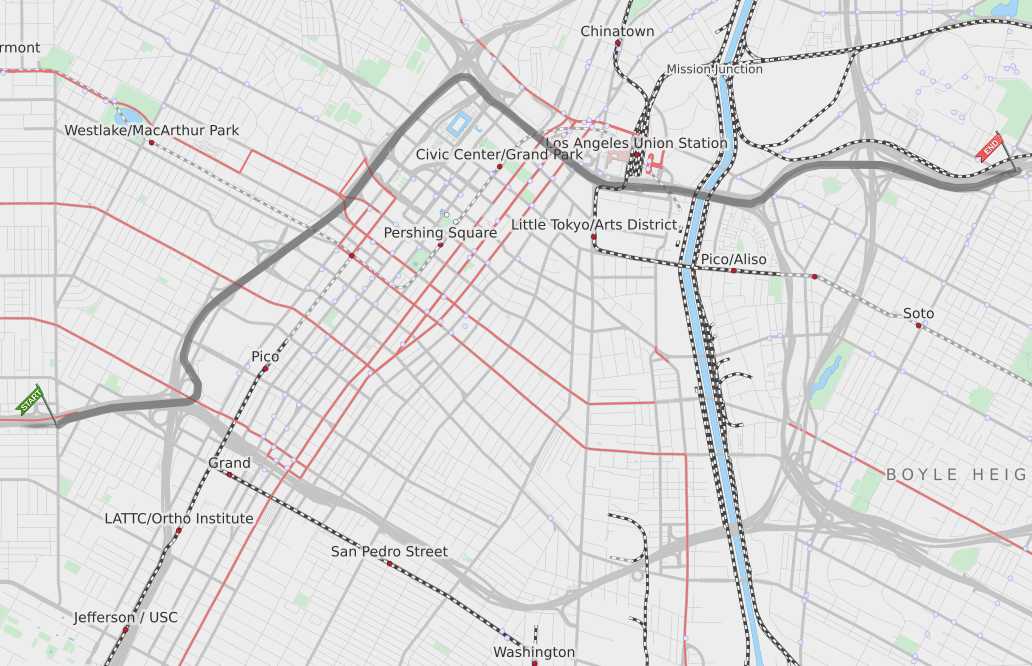}%
\label{a}}
\hfil
\subfloat[]{\includegraphics[width=2.1in]{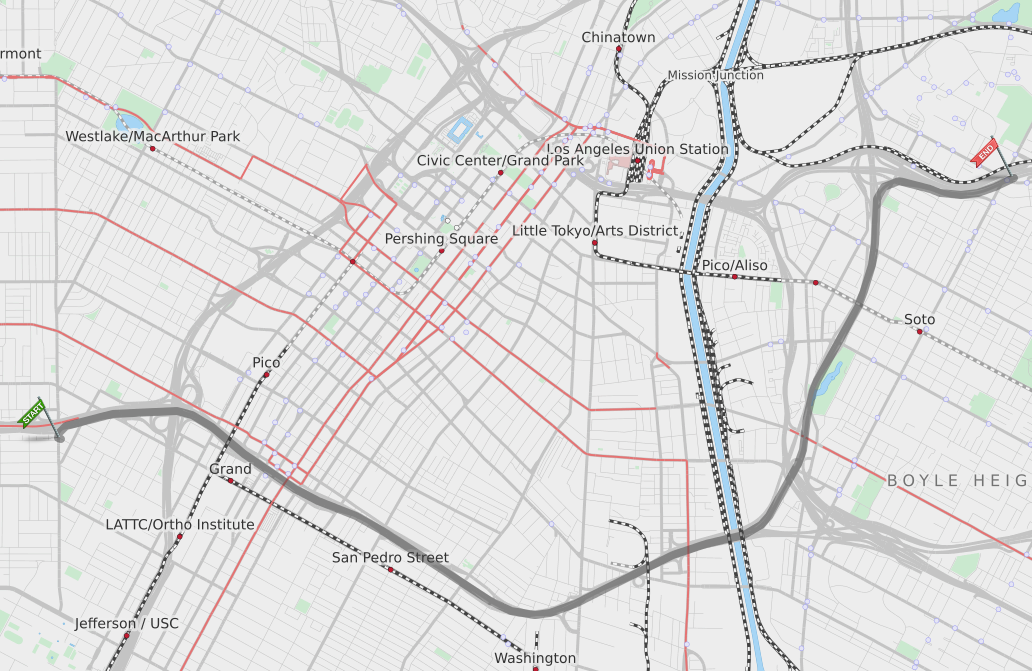}%
\label{b}}
\caption{Reroute occurs due to a conditional undercount fault. (a) Normal. (b) Fault. (Green flag is the source and red flag is the destination.) }
\label{fig:sim}
\end{figure}

We simulate a route planning scenario in OTP, where the edge costs (i.e., travel times) are updated using our traffic data. For a source and destination as shown in Figure~\ref{a}, we consider 1000 queries made on September 15, from 9:00\,am to 10:00\,am. 
Figure~\ref{a} shows the shortest route when a particular sensor (i.e., VDS 774685) is healthy, and Figure~\ref{b} shows the shortest route when the same sensor has a conditional undercount fault.
Note that if the fault remains undetected (i.e., false negative), a suboptimal route (Figure~\ref{b}) will be selected instead of the optimal route (Figure~\ref{a}). 
In another scenario, assume an alarm is triggered under normal operation (i.e., false positive). This means that the predicted value is used for route planning instead of the accurate measurement value, which depending on the prediction accuracy, may result in a suboptimal route planning solution.

\begin{figure}
	\centering
	\includegraphics[width=.85\linewidth]{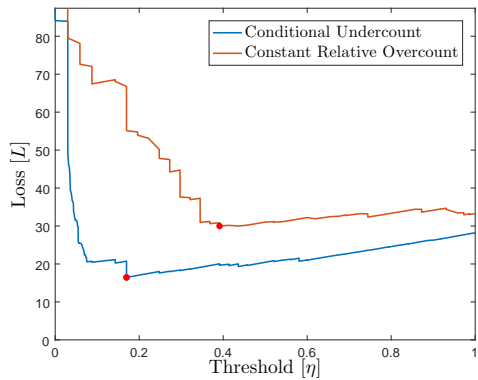}
	\caption{Loss as a function of detection threshold.}
	\label{fig:loss}
\end{figure}

We use Algorithm~\ref{algo} to find near-optimal thresholds that minimize losses due to FPs and FNs. 
We assume that for each sensor, the probability of fault is $p_f = 0.05$. For the previously considered sensor, at $k = 1$ (i.e., from 9:00\;am to 9:05\;am), the loss value \eqref{lossall} as a function of the threshold is shown in Figure~\ref{fig:loss}. In this case, Algorithm~\ref{algo} finds the optimal thresholds. For the Conditional Undercount, the optimal threshold and the minimum loss are $\eta = 0.17$ and $L = 16.2$, whereas for the Constant Relative Overcount, the optimal threshold and the minimum loss are $\eta = 0.39$ and $L = 30.0$.

Further, Table \ref{table:critical} shows the average optimal loss for some sensors, i.e., $\frac{1}{T}\sum_{k=1}^{T} L_s(\eta^*_s(k),Q(k))$. 
As a baseline, we also compute the minimum loss when the thresholds have static values at all the timesteps. That is, for all $k$, we assign $\eta_s(k) = \eta^\ast_s$, where $\eta_s^\ast \in \argmin_{\eta_s} \sum_k L_s(\eta_s, Q)$. We observe that our method achieves significantly smaller losses compared the static case.
The loss values can also be used to identify the set of $\delta$-critical sensors. For example, $50.0$-critical sensors are made bold in the table. 

\begin{table}
	\caption{Average Optimal Losses}
	\label{table:critical}
	\begin{tabular}{>{\centering\arraybackslash}p{1.2cm}>{\centering\arraybackslash}p{1.2cm}>{\centering\arraybackslash}p{1.2cm}>{\centering\arraybackslash}p{1.2cm}>{\centering\arraybackslash}p{1.2cm}}
		\toprule
		\multirow{2}*{Sensor ID} & \multicolumn{2}{c}{Cond. Undercount} & \multicolumn{2}{c}{Cons. Rel. Overcount} \\
		\cmidrule(r{4pt}){2-3} \cmidrule(l){4-5}
		& Optimal & Static & Optimal & Static \\
		\midrule
		774685 & 16.2  & 31.2 & 30.0 &  38.1\\ 
		774672 & 18.0  & 27.6 &  22.1 & 36.7\\ 
		772501 & 15.6 & 24.3 & 12.8 & 19.2\\
		763453  & \textbf{51.8} & 74.3 & \textbf{57.5} & 80.9\\ 
		737158  & 43.0 & 59.6 & \textbf{54.8} & 71.4\\ 
		\bottomrule 
	\end{tabular} 
\end{table}

\section{Related Work}
\label{sec:related}


There are many papers that study traffic prediction. The work in \cite{liebig2017dynamic} uses multivariate kernel regression models to predict traffic flow in a network, considering route planning as the application. In \cite{du2012adaptive}, the paper provides a travel time prediction algorithm in a small scale simulated network. The work in \cite{sun2012network} constructs robust algorithms for short-term traffic flow prediction. Finally, in \cite{kamarianakis2005modeling}, classical time series approaches are used for short-term speed prediction in a network.

The problem of anomaly detection of traffic sensors is reviewed in \cite{lu2008faulty}. The paper categorizes different methods into the three levels of macroscopic, mesoscopic, and microscopic, and provides practical guidelines for anomaly detection. 
The work in \cite{zygourastowards} presents three methods to detect faulty traffic measurements. The methods are based on Pearson's correlation, cross-correlation, and multivariate ARIMA. 
Finally, the work in \cite{robinson2006development} presents a test, which is based on the relationship between flows at adjacent sensors to detect faulty loop detectors. Nevertheless, since previous papers use static thresholds, their methods result in high losses due to FPs and FNs.

In our previous work, we have considered the problem of optimal parameter selection for anomaly detection. The problem of finding optimal thresholds for intrusion detectors is studied in~\cite{laszka2016optimal}. The paper shows that computing optimal attacks and defenses is computationally expensive, and proposes heuristic algorithms for computing near-optimal strategies. Further, the work in \cite{ghafouri2016optimal} studies the problem of finding optimal thresholds for anomaly-based detectors implemented in dynamical systems in the face of strategic attacks. The paper provides algorithms to compute optimal thresholds that minimize losses considering best-response attacks.

\section{Conclusions}
We studied the problem of finding optimal detection parameters for anomaly detection of traffic sensors, considering route planning as application. We constructed a predictor using Gaussian processes, which was then used for anomaly detection. We 
studied how to find the optimal detection parameters, which minimize losses due to FP and FN errors. 
We also characterized critical sensors, whose failure can have high impact on the traffic application. We implemented our method and evaluated it numerically using a route-planning platform.
Our evaluations indicated that the proposed detection method successfully minimizes the performance losses.

\begin{acks}
The work is supported by the \grantsponsor{}{National Science Foundation}{} (\grantnum{}{CNS-1238959}, \grantnum{}{CNS-1647015}), \grantsponsor{}{the Air Force Research Laboratory}{} (\grantnum{}{FA 8750-14-2-0180}), and \grantsponsor{}{the National Institute of Standards and Technology}{} (\grantnum{}{70NANB15H263})
\end{acks}


\balance
\bibliographystyle{ACM-Reference-Format}
\bibliography{reference} 

\end{document}